\documentclass[letterpaper, 10 pt, conference]{ieeeconf}  % Comment this line out if you need a4paper

\IEEEoverridecommandlockouts                              % This command is only needed if 
\usepackage{amsmath} % assumes amsmath package installed
\usepackage{amssymb}  % assumes amsmath package installed
\usepackage{mathrsfs}%用于产生一种数学用的花体字  
\usepackage{amsmath}%数学符号  
\usepackage{bm}%专门处理数学粗体的bm宏包
\usepackage[cmintegrals]{newtxmath}%高质量数学字体/黑体 
\usepackage{booktabs}
\usepackage{multirow}
\usepackage{cite}
\usepackage{color}
\usepackage{graphicx}
\usepackage{subfigure}
\usepackage{threeparttable}
\usepackage{balance}
\usepackage{babel}

\usepackage{hyperref}
\hypersetup{hypertex=true,
            colorlinks=true,
            linkcolor=black,
            anchorcolor=black,
            citecolor=black}

\title{\LARGE \bf
Fast and Robust Lidar-Inertial Odometry by Tightly-Coupled Iterated Kalman Smoother and Robocentric Voxels
}

\author{Jun Liu$^*$, Yunzhou Zhang, Xiaoyu Zhao$^*$, Zhengnan He$^{1}$
	\thanks{$^{1}$Jun Liu, Yunzhou Zhang, Xiaoyu Zhao, Zhengnan He are with College of Information Science and Engineering, Northeastern University, Shenyang 110819, China (Email: {\tt\small zhangyunzhou@mail.neu.edu.cn}).}
  \thanks{$^*$These authors contributed equally to this work and should be considered co-first authors.}
}

\begin{document}

\maketitle
\thispagestyle{empty}
\pagestyle{empty}
\bibliographystyle{IEEEtran} %换风格
% \graphicspath{ {image/} }  %image为文件夹名，可以在左侧自己创建文件夹

%%%%%%%%%%%%%%%%%%%%%%%%%%%%%%%%%%%%%%%%%%%%%%%%%%%%%%%%%%%%%%%%%%%%%%%%%%%%%%%%
\begin{abstract}
This paper presents a fast lidar-inertial odometry (LIO) that is robust to aggressive motion. To achieve robust tracking in aggressive motion scenes, we exploit the continuous scanning property of lidar to adaptively divide the full scan into multiple partial scans (named sub-frames) according to the motion intensity. And to avoid the degradation of sub-frames resulting from insufficient constraints, we propose a robust state estimation method based on a tightly-coupled iterated error state Kalman smoother (ESKS) framework. Furthermore, we propose a robocentric voxel map (RC-Vox) to improve the system's efficiency. The RC-Vox allows efficient maintenance of map points and k nearest neighbor (k-NN) queries by mapping local map points into a fixed-size, two-layer 3D array structure. Extensive experiments are conducted on 27 sequences from 4 public datasets and our own dataset. The results show that our system can achieve stable tracking in aggressive motion scenes (angular velocity up to 21.8 rad/s) that cannot be handled by other state-of-the-art methods, while our system can achieve competitive performance with these methods in general scenes.
Furthermore, thanks to the RC-Vox, our system is much faster than the most efficient LIO system currently published.
\end{abstract}
%%%%%%%%%%%%%%%%%%%%%%%%%%%%%%%%%%%%%%%%%%%%%%%%%%%%%%%%%%%%%%%%%%%%%%%%%%%%%%%%

\section{INTRODUCTION}

\begin{figure}       %不带*单栏，带*双栏 
    \centering
    \includegraphics[scale=0.56]{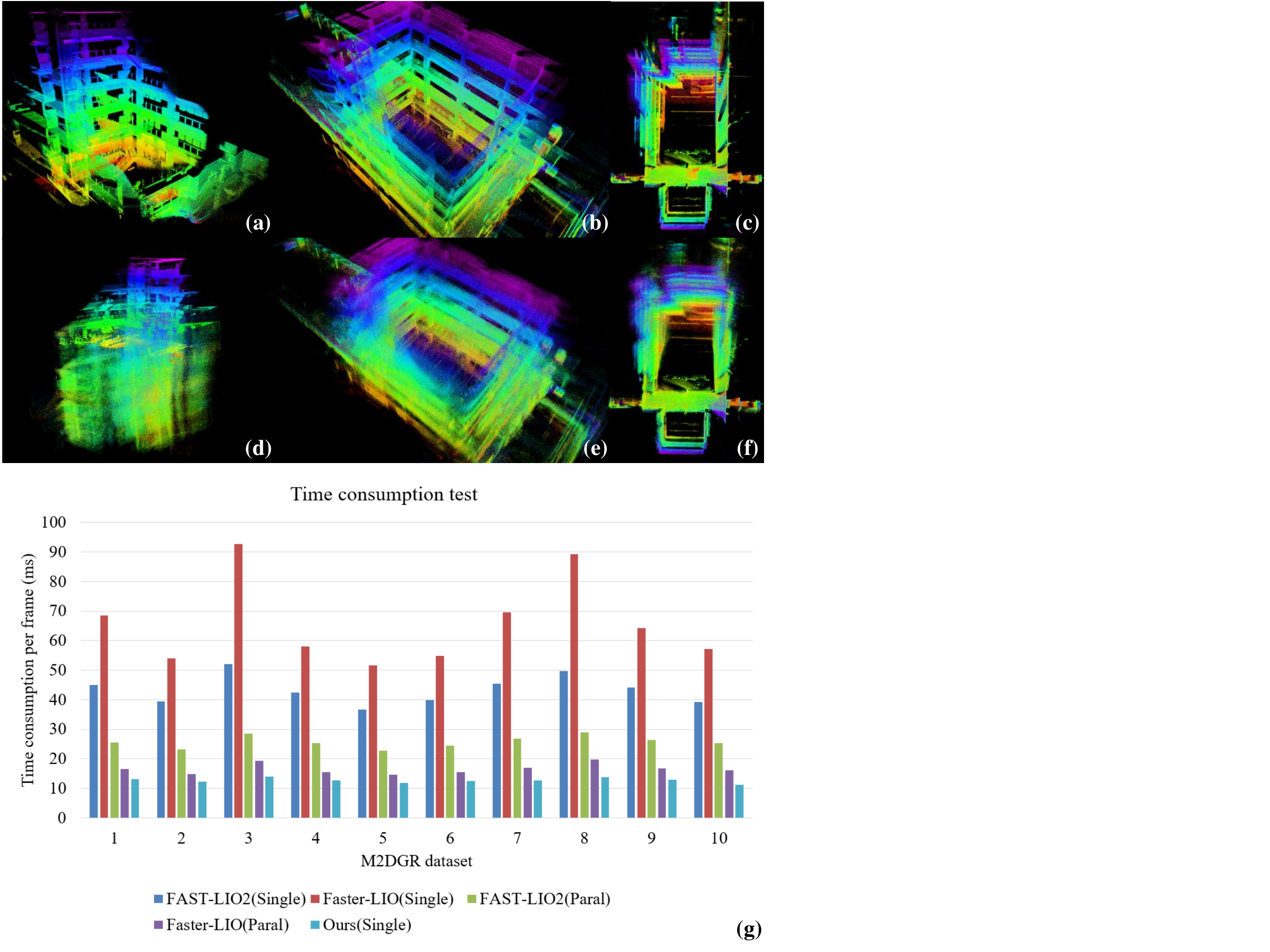}  %scale缩放比例，Fig.jpg文件名
    \caption{The aggressive motion test on our own dataset (compared with FAST-LIO2) and the speed test on the M2DGR dataset (compared with FAST-LIO2 and Faster-LIO). (a)-(c) and (d)-(f) are the point cloud reconstructed by our system and FAST-LIO2 respectively on \emph{indoor\_01-03} of our own dataset. (g) is the system speed test on \emph{street\_1-10} of the M2DGR dataset, where \emph{(paral)} means that the system is parallelized, and \emph{(single)} means that the system is running in single thread mode.}   % 图片名称
    \label{fig:1}
    \vspace{-0.4cm}
\end{figure}

Simultaneous localization and mapping (SLAM) is a crucial technique for autonomous mobile robots to navigate in an unknown environment. SLAM can be divided into vision-based SLAM and lidar-based SLAM, among which lidar-based SLAM is widely used because it can provide fast, accurate, and dense 3D reconstruction. In lidar-based SLAM, the odometry module significantly impacts the system's overall performance, so this paper only focuses on the odometry part.

The performance of the LO/LIO system is mainly determined by accuracy, robustness, and efficiency. In terms of accuracy, some approaches improve registration accuracy through more accurate probabilistic map representation \cite{yuan2022efficient}, more refined lidar points segmentation \cite{pan2021mulls}\cite{chen2021psf}, or selecting more useful lidar points for registration \cite{deschaud2018imls}. In addition, some approaches improve the accuracy by reducing or avoiding motion distortion of the lidar points \cite{dellenbach2022ct}\cite{lv2021clins}.

As for the robustness, it mainly focuses on the ability of the system to deal with various corner cases, such as scene degradation, aggressive motion, and so on. In degraded scenes, only using lidar cannot solve this problem well. Some approaches \cite{shan2021lvi, lin2021r, lin2022r, zheng2022fast} achieve stable tracking in lidar-degraded scenes by fusing vision. In aggressive motion scenes, the limitation of the IMU frequency prevents it from capturing the full motion information, resulting in inaccurate motion compensation of lidar points and poor initial pose of registration. Although some approaches can cope with a certain degree of aggressive motion \cite{lv2021clins}\cite{qu2022llol}, there is not yet a reliable solution to this problem. In this paper, we reduce the IMU integration error by adaptively dividing the full lidar scan into multiple sub-frames according to the motion intensity. To deal with the possible insufficient constraints of the sub-frames, we propose a tightly-coupled iterated ESKS framework for state estimation, which smoothes the degraded sub-frame pose by using both the historical and future constraints. Using the above approach, we achieve robust tracking in aggressive motion scenes.

For the system efficiency, the chosen spatial data structure, the state estimator formation, and the residual metric for matching affect the computation time cost significantly \cite{bai2022faster}. Among these factors, the spatial data structure is the most time-consuming module of the system. Hence, a better spatial data structure is crucial to keep the system lightweight and efficient. Compared with the tree structure, the hash voxel's insertion, deletion, and query operation have lower time complexity. However, due to the complex calculations in the hash function and the inevitable hash collisions, the hash table does not really achieve the same efficient access as the array structure. Although the array structure has a fast access performance, its rapidity is achieved through a large memory consumption, which makes it rarely used for the large-scale point cloud representation. In this paper, we make it possible to use the array structure in the LIO system by designing a fixed-size two-layer 3D array structure, local map points within the lidar range are maintained by this fixed-size structure with a modulo operation. In addition, a method for refining the k-NN process is proposed, which further improves the system's efficiency.

In summary, the contributions of this paper are as follows:

\begin{itemize}

\item We make the LIO system robust to aggressive motion via an adaptive lidar points division. In addition, we propose an iterated ESKS-based state estimation framework to avoid the degradation of the sub-frame and guarantee the smoothness of the trajectory.

\item We propose a robocentric voxel map structure, called RC-Vox, which enables fast incremental mapping through a fixed-size, two-layer 3D array structure. Moreover, a refined k-NN search strategy is proposed to accelerate the point cloud registration.

\item We verify our system on 4 public datasets and our own dataset. Experiments show that our system achieves competitive performance against the state-of-the-art LIO system in general scenes and is robust to aggressive motion. Moreover, the single-thread version of our system is even faster than the parallel accelerated version of the most efficient LIO system.

\end{itemize}

\section{RELATED WORK}
In this section, we review the state-of-the-art LIO systems and the spatial data structures commonly employed in LIO systems.

\subsection{Lidar-Inertial Odometry}
In recent years, some outstanding LIO systems have emerged, we only discuss the parts that are relatively robust to aggressive motion. FAST-LIO2 \cite{xu2022fast} downsamples the original lidar points and performs scan-to-map registration directly in an iterated error state Kalman filter framework. This framework follows the new Kalman gain calculation method proposed in FAST-LIO \cite{xu2021fast} to speed up the system. Although FAST-LIO2 is assisted by an IMU, it does not handle the overly aggressive motion. The continuous-time method like CLINS \cite{lv2021clins} is able to ignore the motion compensation error caused by the aggressive motion through the non-rigid registration strategy. However, this requires more control points to fit the violently varying trajectory, which greatly affects the system's efficiency. LLOL \cite{qu2022llol} treats spinning lidar as a streaming sensor by dividing the lidar points. The problem of insufficient constraints on partial scans is avoided by using a circular buffer to ensure a full scan of measurements are used for registration. Because it needs to estimate the pose by the amount of data in the complete frame when processing each partial scan, the system becomes less efficient as the number of partial scans increases. Therefore, it is also unable to handle larger degrees of aggressive motion.

\subsection{Spatial Data Structure}
Many approaches \cite{qin2020lins,shan2020lio, xu2021fast, lv2021clins, li2021towards} use the kd-tree for map maintenance, which can achieve fast k-NN search. Nevertheless, the whole k-d tree has to be reconstructed when new points are added or old points are deleted, which greatly reduces the efficiency of the system. The ikd-Tree structure proposed by FAST-LIO2 \cite{xu2022fast} greatly reduces the time consumption of map update compared with the static k-d tree structure. However, partial reconstruction is still required to achieve balance when adding or deleting points that lead to an imbalance in the tree. Compared with the k-d tree structure, the hash voxel is more efficient for map maintenance and has a faster query speed. The iVox structure proposed in Faster-LIO \cite{bai2022faster} organizes voxels through a hash table and a LRU cache, whose parallel accelerated version can achieve a performance that completely surpasses the performance of the ikd-Tree. However, due to the existence of collisions, the hash structure are not as fast to access as the array structure, which leaves the potential for the voxel map to be further improved.

\section{System Overview}
\begin{figure}
    \vspace{0.2cm}
    \centering
    \includegraphics[scale=0.387]{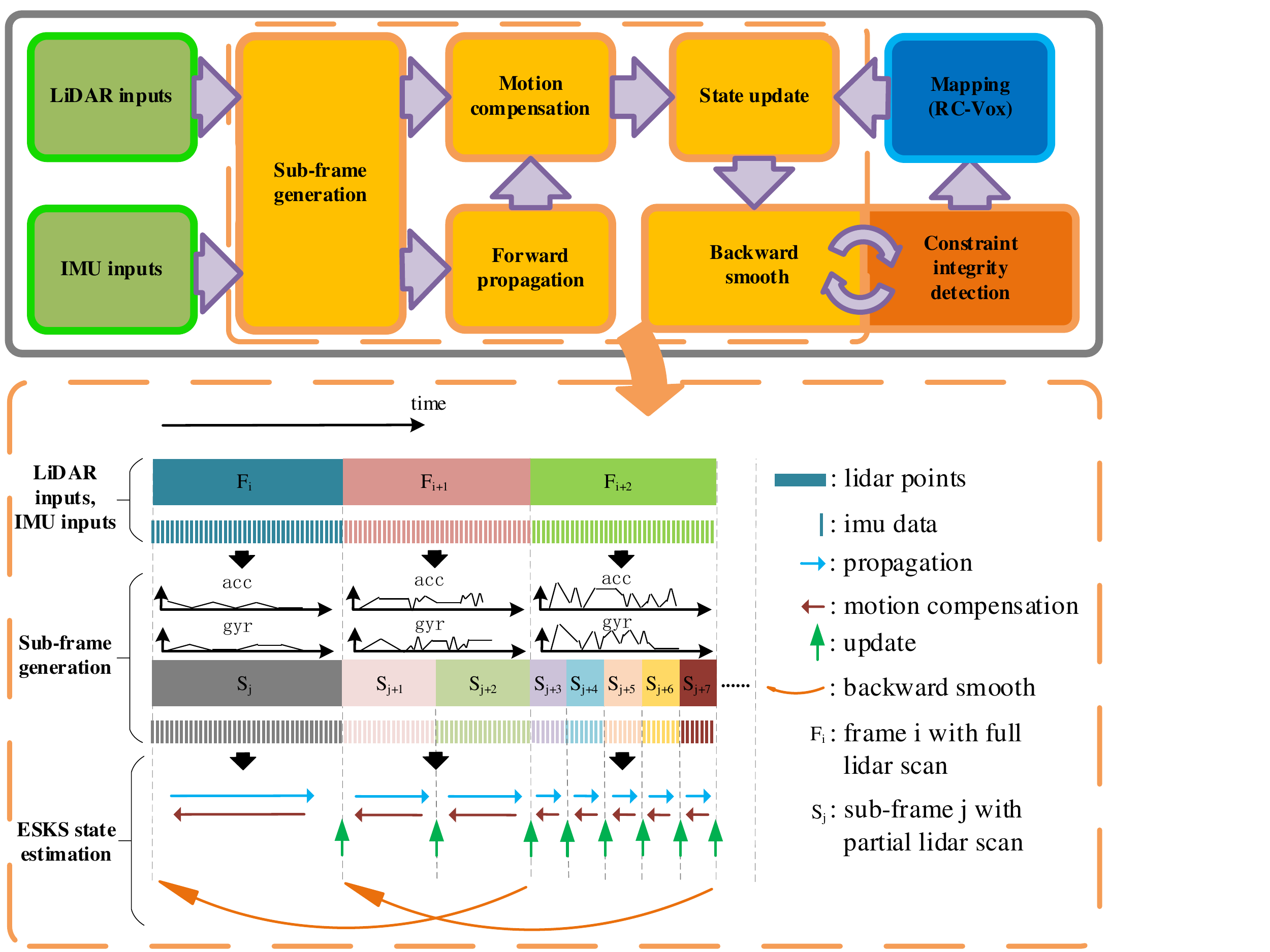}  %scale缩放比例，Fig.jpg文件名
    \caption{System overview.}   % 图片名称
    \label{fig:2}
    \vspace{-0.3cm}
\end{figure}

The pipeline of the system is shown in Fig. \ref{fig:2}. The lidar scan is fed into the system together with the IMU measurements. By detecting the sensor motion intensity, the sub-frame generation module first determines how many sub-frames the full lidar scan will be divided into and then generate multiple sub-frames (see \ref{sec:Sub-frame Generation}). Then, we performs forward propagation of the IMU measurements and motion compensation (see \ref{sec:Propagation}), after which we uses the undistorted lidar points to perform state update with a direct method (see \ref{sec:Update}). Backward smooth is then performed using sub-frames corresponding to at least three full scans (see \ref{sec:Backward Smooth}). Finally, the smoothed poses are sent to the constraint integrity detection module to check if the constraints are sufficient (see \ref{sec:Constraint Integrity Detection}), and the pose with no degradation is used to register the lidar points to the RC-Vox (see \ref{sec:RC-Vox: Robocentric Voxel Map}).

\section{State Estimation}
\subsection{Sub-frame Generation}
\label{sec:Sub-frame Generation}
The system adaptively determines the number of sub-frames to be generated based on the motion intensity. Although generating more sub-frames can further reduce the impact of IMU integration error on motion compensation theoretically, the presence of dynamic objects in the environment and cluttered points will greatly reduce the system stability if only a few number of points are used for registration. Therefore, it only makes sense to divide full scan into more sub-frames in aggressive motion scenes. Based on the above reasons, we count the standard deviation of acceleration and angular velocity for different motion intensities and simply assume that the number of sub-frames is proportional to the standard deviation value. Assume that $\sigma _{acc\_max} $, $\sigma _{gyr\_max}$  are the maximum standard deviations of acceleration and angular velocity in all directions derived from statistics respectively, $\sigma _{acc\_curr} $, $\sigma _{gyr\_curr}$ are the standard deviations of acceleration and angular velocity corresponding to the current scan, $n_{max}$ is the maximum number of sub-frames allowed to be divided from a full lidar scan, {$n_{max}$ depends on the total number of points in each frame, and we use different $n_{max}$ for lidar sensors from different manufacturers.} then the number of sub-frames to be divided $n_{curr}$ is:
\begin{equation}
n_{curr} = \mathrm{ceil}\left ( n_{max} * max\left ( \frac{\sigma _{acc\_curr}}{\sigma _{acc\_max}}, \frac{\sigma _{gyr\_curr}}{\sigma _{gyr\_max}} \right )  \right ) 
\end{equation}
where $\mathrm{ceil}()$ stands for rounding up.

\subsection{Propagation} 
\label{sec:Propagation}

The system nominal state $\bm{x} $, error state $\bm{\delta x}$, IMU input $\bm{u} $ and noise $\bm{w} $ are defined as follows:
\begin{equation}
\begin{aligned}
& 
\mathcal{M}\triangleq\mathbb{R}^{6}\times SO(3)\times \mathbb{R}^{9}   
\\ & 
\bm{x}\triangleq\begin{bmatrix}
 ^{G}\bm{p}_{b}^{T} & ^{G}\bm{v}_{b}^{T} & ^{G}\bm{q}_{b}^{T} & \bm{b}_{a}^{T} & \bm{b}_{g}^{T} & ^{G}\bm{g}^{T}
\end{bmatrix}^{T} \in \mathcal{M}
\\ & 
\delta \bm{x}\triangleq\begin{bmatrix}
 ^{G}\delta \bm{p}_{b}^{T} & ^{G}\delta \bm{v}_{b}^{T} & ^{G}\delta \bm{\theta }_{b}^{T} & \delta \bm{b}_{a}^{T} & \delta \bm{b}_{g}^{T} & ^{G}\delta \bm{g}^{T}
\end{bmatrix}^{T} \in \mathbb{R}^{18}
\\ & 
\bm{u}\triangleq\begin{bmatrix}
 \bm{a}_{m}^{T} & \bm{\omega }_{m}^{T}
\end{bmatrix}^{T}
\\ & 
\bm{w}\triangleq\begin{bmatrix}
 \bm{n}_{a}^{T} & \bm{n}_{\omega}^{T} & \bm{n}_{ba}^{T} & \bm{n}_{b\omega}^{T}
\end{bmatrix}^{T}
\end{aligned}
\end{equation}
where $^{G}\bm{p}_{b}$, $^{G}\bm{v}_{b}$, $^{G}\bm{q}_{b}$ are the IMU position, velocity, attitude in the global frame, $\bm{b}_{a}$, $\bm{b}_{g}$ are the IMU biases and $^{G}\bm{g}$ is the gravity in the global frame. $ \bm{a}_{m}$, $\bm{\omega }_{m}$ are IMU measurements. $\bm{n}_{a}$, $\bm{n}_{\omega}$ are the measurement noise of $ \bm{a}_{m}$, $\bm{\omega}_{m}$ and $\bm{n}_{ba}$, $\bm{n}_{b\omega}$ are the random walk noise of IMU biases.
Using the IMU measurements as input, the nominal state values are predicted using a discrete-time propagation model:
\begin{equation}
\bm{x}_{k+1}=f(\bm{x}_{k},\bm{u}_{k}) 
\end{equation}
with a linearized discrete-time error state transfer equation \eqref{4}, the covariance matrix of the error state is forward propagated as equation \eqref{5}. 
\begin{equation}
\label{4}
\delta \bm{x}_{k+1} = (I + \bm{F}_{k}\Delta t)\delta \bm{x}_{k} + (\bm{C}_{k}\Delta t)\bm{w}
\end{equation}
\begin{equation}
\label{5}
\bm{P}_{k+1} = (I + \bm{F}_{k}\Delta t)\bm{P}_{k}(I + \bm{F}_{k}\Delta t)^{T} + (\bm{C}_{k}\Delta t) \bm{Q}_{k} (\bm{C}_{k}\Delta t)^{T}
\end{equation}
For the detailed forms of $f(\bm{x}_{k},\bm{u}_{k})$, $\bm{F}_{k}$, $\bm{C}_{k}$ and $\bm{Q}_{k}$, readers can refer to \cite{sola2017quaternion}. In addition, the result of the forward propagation are used for motion compensation and registration.

\subsection{Update} 
\label{sec:Update}
We follow the direct method of FAST-LIO2 \cite{xu2022fast} by downsampling the sub-frame lidar points and searching the 5 nearest neighbors for each point (see \ref{sec:K-NN Search}), then constructing point-to-plane residuals after fitting the local plane patches:
\begin{equation}
h_{j}\left ( \bm{x}_{k}^{\kappa } \right ) =^{G}\bm{n}_{j}^{T}\left ( ^{G}\bm{q}_{b_{k}}^{\kappa } \left ( ^{b}q_{l}{^{l}\bm{p}_{j}}+^{b}\bm{p}_{l} \right ) +  ^{G}\bm{p}_{b_{k}}^{\kappa } - ^{G}\bm{c}_{j}\right ) 
\end{equation}
where $h_{j}\left ( \bm{x}_{k}^{\kappa } \right )$ is the j-th row of $h\left (\bm{x}_{k}^{\kappa} \right )$, $^{b}\bm{q}_{l}$ and $^{b}\bm{p}_{l}$ are the pre-calibrated extrinsic between lidar and IMU, $^{G}\bm{n}_{j}$ and $^{G}\bm{c}_{j}$ are normal vector and centroid of the j-th plane patch respectively, $^{l}\bm{p}_{j}$ is the corresponding j-th lidar point, $\kappa$ means the $\kappa$-th iteration.

Using the point-to-plane residuals as measurements, data association and error state update operations are performed iteratively \cite{xu2022fast}:
\begin{equation}
\bm{K} = (\bm{H}^{T}\bm{R}^{-1}\bm{H}+\bm{P}_{k}^{-1})^{-1}\bm{H}^{T}\bm{R}^{-1}
\end{equation}
\begin{equation}
\bm{x}_{k}^{\kappa +1} = \bm{x}_{k}^{\kappa} + \bm{K}\left ( \bm{0}-h(\bm{x}_{k}^{\kappa}) \right ) - \left ( \bm{I}-\bm{K}\bm{H} \right ) \left (\bm{x}_{k}^{\kappa} -\bm{x}_{k}^{0} \right )
\end{equation}
During the above iterations, we have simultaneously performed correction of the nominal state and reset of the error state. When the iterations converge, the state covariance is updated with the following equation:
\begin{equation}
\bm{P}_{k} = \left ( \bm{I}-\bm{K}\bm{H} \right )  \bm{P}_{k}
\end{equation}

\subsection{Backward Smooth} 
\label{sec:Backward Smooth}

Since the original lidar scan is divided into sub-frames, which may be degraded resulting from insufficient constraints. Once the system has finished processing all the sub-frames corresponding to the current scan, we perform a backward smooth on the currently existing sub-frames in the system:
\begin{equation}
\label{10}
\bm{x}_{k+1}^{-} = f(\bm{x}_{k},\bm{u}_k)
\end{equation}
\begin{equation}
\label{11}
\begin{split}
\bm{P}_{k+1}^{-} = \left ( \bm{I} + \bm{F}_{k}\Delta t \right )  \bm{P}_{k} \left ( \bm{I} + \bm{F}_{k}\Delta t \right )^{T} \\ +  \left (\bm{C}_{k}\Delta t \right )  \bm{Q}_{k} \left (\bm{C}_{k}\Delta t \right )^{T}
\end{split}
\end{equation}
\vspace{0.01cm}
\begin{equation}
\label{12}
\bm{G}_{k} = \bm{P}_{k}\left ( \bm{I} + \bm{F}_{k}\Delta t \right )^{T}\left [ \bm{P}_{k+1}^{-} \right ]^{-1} 
\end{equation}
\begin{equation}
\label{13}
\bm{x}_{k}^{s} = \bm{x}_{k} + \bm{G}_{k}\left [ \bm{x}_{k+1}^{s} - \bm{x}_{k+1}^{-} \right ]
\end{equation}
\begin{equation}
\label{14}
\bm{P}_{k}^{s} = \bm{P}_{k} + \bm{G}_{k}\left [ \bm{P}_{k+1}^{s} - \bm{P}_{k+1}^{-} \right ]\bm{G}_{k}^{T}
\end{equation}
where $f(\bm{x}_{k},\bm{u}_{k})$, $\bm{F}_{k}$, $\bm{C}_{k}$ and $\bm{Q}_{k}$ are the same as in propagation, $\bm{x}_{k}^{s}$ and $\bm{P}_{k}^{s}$ are the results of the backward smooth which is started from the last sub-frame at time $T$, with $\bm{x}_{T}^{s} = \bm{x}_{T}$ and $\bm{P}_{T}^{s} = \bm{P}_{T}$. A detailed introduction of the Kalman smoother can be found in \cite{sarkka2013bayesian}. In order to apply Kalman smoother to the error state, we make some modifications to the backward smooth part (as shown in equations \eqref{10}-\eqref{14}), the validity of these equations is explained below.

The error state Kalman filter is equivalent to using the extended Kalman filter to manipulate the error state, then using the error state to correct the nominal state and reset the error state after each update. However, the reset operation makes it unable to apply the extended Kalman smoother (EKS) directly. To operate the error state in EKS, we temporarily cancel the reset operation on the error state and perform forward propagation, update and backward smooth on the unreseted error state. Meanwhile, we calculated $\bm{F}_{k}$ with the sum of the unreseted error state and the uncorrected nominal state.
Under the assumption that the error state always varies around 0, EKS can be applied to the error state. Assume that $\delta\bm{x}_{k}^{p}$, $\delta\bm{x}_{k}^{u}$, $\delta\bm{x}_{k}^{s}$ are the error states after forward propagation, update and backward smooth at time k respectively, $\delta\bm{x}_{k}^{\Delta u}$ and $\delta\bm{x}_{k}^{\Delta s}$ are the corrected parts of update and backward smooth, $\bm{x}_{k}$ is the uncorrected nominal state, the backward smooth part of the EKS is slightly different from equations \eqref{10}-\eqref{14}, where equations \eqref{10} and \eqref{13} are replaced with equations \eqref{15} and \eqref{16}.
\begin{equation}
\label{15}
\delta \bm{x}_{k+1}^{p} = \left ( 1 + F_{k}\Delta t \right )\delta \bm{x}_{k}^{u}
\end{equation}
\begin{equation}
\label{16}
\delta \bm{x}_{k}^{s} = \delta \bm{x}_{k}^{u} + \bm{G}_{k}\left [ \delta \bm{x}_{k+1}^{s} - \delta \bm{x}_{k+1}^{p} \right ] 
\end{equation}

Then, suppose the following equation holds approximately:
\begin{equation}
\begin{aligned}
\bm{x}_{k+1}^{s} &= \bm{x}_{k+1} + \delta \bm{x}_{k+1}^{s} \\
&= \bm{x}_{k+1}+\delta \bm{x}_{k+1}^{u}+\delta \bm{x}_{k+1}^{\Delta s} \\
&= \bm{x}_{k+1}+\delta \bm{x}_{k+1}^{p}+\delta \bm{x}_{k+1}^{\Delta u}+\delta \bm{x}_{k+1}^{\Delta s} \\
&= f\left ( \bm{x}_{k},\bm{u}_k \right ) + \left ( 1 + F_{k}\Delta t \right )\delta \bm{x}_{k}^{u} +\delta \bm{x}_{k+1}^{\Delta u}+\delta \bm{x}_{k+1}^{\Delta s} \\
&\approx  f\left ( \bm{x}_{k}+\delta \bm{x}_{k}^{u},\bm{u}_k \right ) +\delta \bm{x}_{k+1}^{\Delta u}+\delta \bm{x}_{k+1}^{\Delta s}
\end{aligned}
\end{equation}
we can obtain the following relationship:
\begin{equation}
\begin{aligned}
\delta \bm{x}_{k+1}^{s} - \delta \bm{x}_{k+1}^{p} &= \left ( \bm{x}_{k+1}^{s} - \bm{x}_{k+1} \right ) - \delta \bm{x}_{k+1}^{p} \\
&= \bm{x}_{k+1}^{s} - \left ( \bm{x}_{k+1} + \delta \bm{x}_{k+1}^{p}\right )  \\
&\approx  \bm{x}_{k+1}^{s} - f\left ( \bm{x}_{k}+\delta \bm{x}_{k}^{u},\bm{u}_k \right )
\end{aligned}
\end{equation}
where $\bm{x}_{k}+\delta \bm{x}_{k}^{u}$ is the corrected nominal state. Then, by replacing $\delta \bm{x}_{k+1}^{s} - \delta \bm{x}_{k+1}^{p}$ in equation \eqref{16} with $\bm{x}_{k+1}^{s} - f\left ( \bm{x}_{k}+\delta \bm{x}_{k}^{u},\bm{u}_k \right )$ and adding uncorrected $\bm{x}_{k}$ to both sides, we can obtain equation \eqref{13}. Then, the reset of the error state and the correction of the nominal state will not affect the use of Kalman smoother anymore.

\subsection{Constraint Integrity Detection}
\label{sec:Constraint Integrity Detection}
% We use the covariance matrix of the system state and the smooth count to check the constraint integrity of the sub-frame. The role of the covariance matrix is obvious. The reason we use the smooth count is that in the non-degradation scenarios, it is generally considered that the constraint information in a full scan is sufficient. Therefore, the constraint integrity of the current sub-frame is guaranteed using its corresponding lidar scan and the two adjacent lidar scans.
We use the covariance matrix of the system state to check the constraint integrity of the sub-frame. For each sub-frame currently being processed, we check it in time-order from old to new. If the minimum eigenvalue of the covariance matrix is greater than a specified threshold, the constraint is considered to be sufficient. Finally, the sub-frames that are no longer degraded are sent to the RC-Vox for incremental mapping.

\section{RC-Vox: Robocentric Voxel Map}
\label{sec:RC-Vox: Robocentric Voxel Map}

In this section, we introduce the proposed RC-Vox, a robocentric voxel map structure. Unlike iVox \cite{bai2022faster} which uses a combination of hash table and LRU cache, RC-Vox implements efficient k-NN search and incremental mapping directly in a fixed-size, two-layer 3D array structure. We start by explaining some terms for a better introduction to RC-Vox:

\begin{itemize}

\item \textbf{Global coordinate system} is the one used in general LIO system, sometimes it is aligned with the pose of the first scan.

\item RC-Vox maintains the \textbf{local map} points in a cube centered on the robot and with twice the lidar range as edge length, which is sufficient for stable lidar points registration. In addition, the local map points are described under the global coordinate system.

\item \textbf{Top-level array (TLA)} and \textbf{bottom-level array (BLA)} together form a two-layer 3D array structure. Each element in the TLA is called a \textbf{grid}, the internal space of each grid is filled by a BLA, and each element in a BLA is called a \textbf{voxel}.

\item A \textbf{TLA index} corresponds to a grid in the TLA.

\item A \textbf{BLA index\textbf} corresponds to a voxel in a BLA.

\end{itemize}

Moreover, we explain some notations that will be used below. $\bm{r}$, $\bm{m}$, $\bm{t}$, and $\bm{b}$ are the global coordinate of the robot, local map origin, TLA origin, and BLA origin respectively. $\bm{\mathrm{IT}}$ and $\bm{\mathrm{IB}}$ are the TLA index and the BLA index respectively. In addition, all of the above are three-dimensional real or integer vectors.

\subsection{Data Structure of RC-Vox}
The local map points are firstly organized by TLA. For memory consumption consideration, the resolution of the TLA cannot be too high, usually at the meter level. To prevent the excessive points present in each voxel, for a grid with map points, a BLA is generated internally to achieve higher resolution requirements. Assume that the lidar range is $l$, the grid size is $g$ and the voxel size is $v$, the TLA can express a range of $2\lambda l\times 2\lambda l\times 2\lambda l$ space $\left(\lambda>1\right)$, the number of grids in each dimension of the TLA is $2\lambda l/g$, and the number of voxels in each dimension of a BLA is $g/v$.

\subsection{Incremental Mapping}
\begin{figure}
    \vspace{0.1cm}
    \centering
    \setlength{\abovecaptionskip}{-0.5cm}
    \includegraphics[scale=0.33]{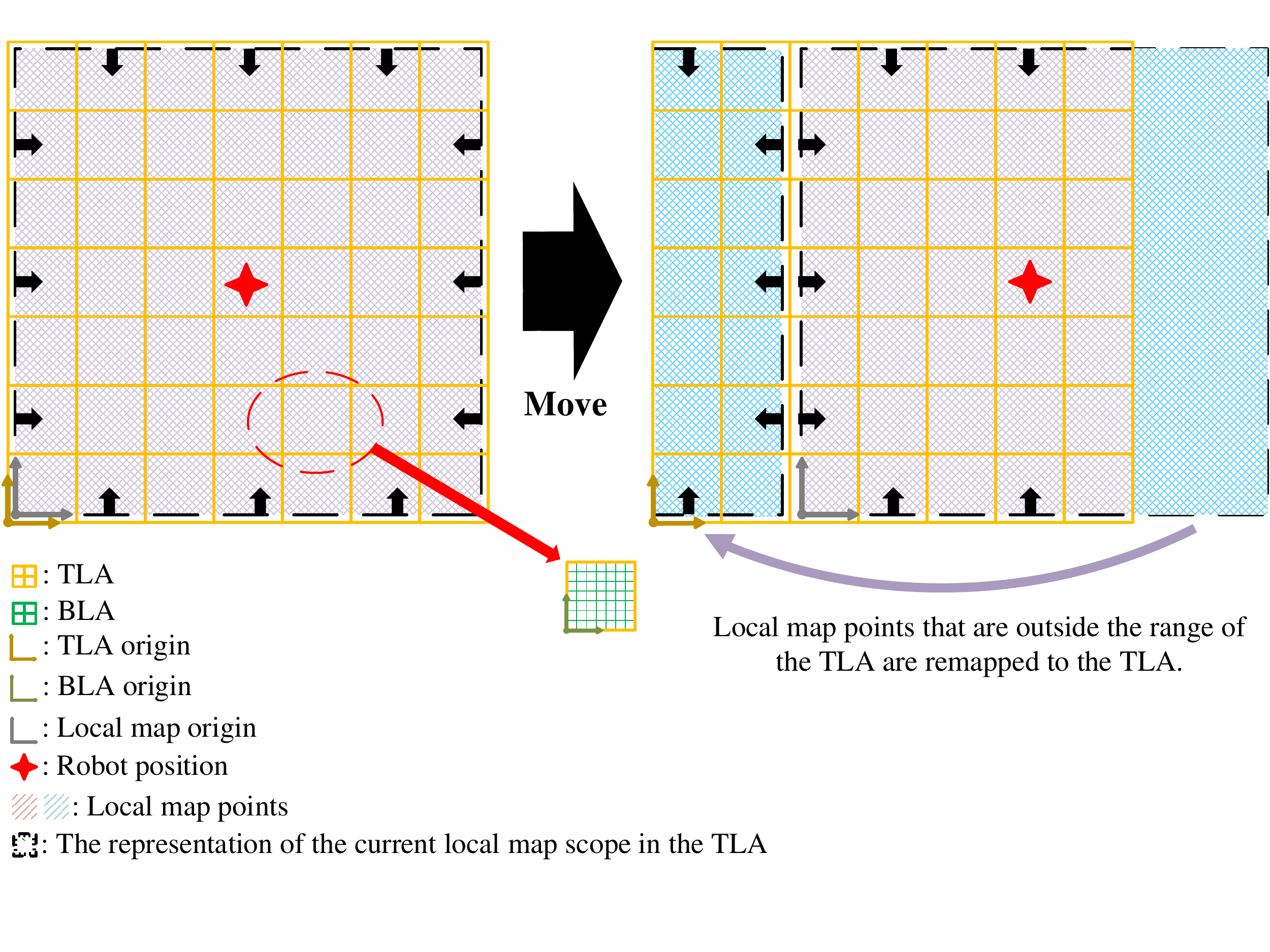}  %scale缩放比例，Fig.jpg文件名
    \caption{2D demonstration of the RC-Vox. The left figure shows the initialization phase, and the right figure shows the process of remapping the local map parts beyond the TLA back to the TLA after the robot moves.}   % 图片名称
    \label{fig:3}
    \vspace{-0.4cm}
\end{figure}

In the system initialization phase, we initialize the voxel map by placing the robot at the center of the TLA, while making the local map coincide with the TLA (as shown in Fig. \ref{fig:3}). Assuming that the global coordinate of the robot is $  \bm{r}_{init}$, the global coordinate of the TLA origin $\bm{t}_{init}$ is:
\begin{equation}
 \bm{t}_{init} = \bm{r}_{init}-\lambda l \bm{I}_{3\times1}
\end{equation}
The local map origin $\bm{m}_{init}$ has the same global coordinate as the TLA origin at this moment and the TLA index of the local map origin $\bm{\mathrm{IT}}_{m_{init}}$ is $\bm{0}_{3\times1}$. It should be noted that the position of the TLA in the global coordinate system is fixed for the whole subsequent mapping period, and a modulo operation is used to map the continually moving local map to the TLA.

With the robot moving, suppose that the current global coordinate of the robot at a certain moment is $\bm{r}_{curr}$, since the local map moves with the robot and is aligned to grids, the global coordinate of the local map origin $\bm{m}_{curr}$ at this moment is:
\begin{equation}
\bm{m}_{curr} = \mathrm{floor}\left ( \left ( \bm{r}_{curr}-\bm{t}_{init} \right ) /g \right ) g + \bm{t}_{init}
\end{equation}
where $\mathrm{floor}()$ stands for rounding down. Meanwhile, the TLA index of the local map origin $\bm{\mathrm{IT}}_{m_{curr}}$ is:

\begin{equation}
\bm{\mathrm{IT}}_{m_{curr}} = \mathrm{floor}\left ( \left ( \bm{r}_{curr}-\bm{t}_{init} \right ) /g \right ) \% \left ( 2\lambda l/g \right )
\end{equation}
For each local map point $\bm{p}$, the TLA index $\bm{\mathrm{IT}}_{p}$ is:
\begin{equation}
\label{22}
\bm{\mathrm{IT}}_{p} = \left (\mathrm{floor}\left ( \left ( \bm{p}-\bm{m}_{curr} \right ) /g \right ) + \bm{\mathrm{IT}}_{m_{curr}}\right )\% \left ( 2\lambda l/g \right )
\end{equation}
It can be seen that the local map is remapped into the TLA by a modulo operation. When we register lidar points to the map, we also use the equation \eqref{22} to find the TLA index corresponding to the current point. If the grid corresponding to this TLA index is not yet used, a BLA is generated in this grid, and the global coordinate of this BLA origin is recorded, where the global coordinate of the origin $\bm{b}_{ori}$ is:
\begin{equation}
\bm{b}_{ori} = \mathrm{floor}\left ( \left ( \bm{p}-\bm{t}_{init} \right ) /g \right ) g + \bm{t}_{init}
\end{equation}
we proceed to calculate its BLA index, and record this point into the corresponding voxel, where the BLA index $\bm{\mathrm{IB}}_{p}$ is:

\begin{equation}
\label{24}
\bm{\mathrm{IB}}_{p} = \mathrm{floor}\left (\left ( \bm{p}-\bm{b}_{ori} \right ) /v)\right )
\end{equation}

As shown in Fig. \ref{fig:3}, the TLA is set to the same size as the local map at the initialization phase, therefore, the remapping operation will not cause problem with overlapping of the local map points in the TLA.
In addition, we only maintain the local map points in a cube, so we can likewise calculate the TLA index of the cube boundary.
When a robot position change leads to the cube moving in the TLA, a portion of the grids will move out of the cube while these grids will re-enter the cube on the other side. At the same time, the contents of these grids are reset, enabling the incremental construction and deletion of the map points.

\subsection{K-NN Search}
\label{sec:K-NN Search}
When conducting lidar points registration, we query the voxels as the approach in equations \eqref{22}, \eqref{24}. In order to perform an accurate k-NN search, not only the voxel corresponding to the current point but also the nearby voxels need to be searched. 
% iVox \cite{bai2022faster} searches its nearby 6(/18/26) voxels additionally and our k-NN search strategy is similar, but with some improvements. 
Since multiple rounds of k-NN search are performed during the lidar points registration, a significant portion of the time consumption of the LIO system is spent on the k-NN search. RC-Vox uses iVox's \cite{bai2022faster} k-NN search strategy with some improvements by moving the process of searching nearby points from the registration phase to the mapping phase. When registering a lidar point to map, in addition to filling the corresponding voxel it belongs to, this point is also recorded in the nearby 6(/18/26) voxels. Since the mapping is incremental, only a small number of points need to be added to the map in each round, so the increase in time consumption due to this operation is nearly negligible. When conducting a k-NN search, we can search only the corresponding voxel and get all the points in its neighbor, then calculate the distance from the current point to each of these points and select the k nearest points with the smallest distance as the result of the k-NN search.

\section{Experiment}
To verify the effectiveness of the proposed method, we select 27 sequences from our own dataset and 4 public datasets (NTU VIRAL \cite{nguyen2022ntu}, M2DGR \cite{yin2021m2dgr}, UrbanNav \cite{hsu2021urbannav}, Newer College \cite{ramezani2020newer, zhang2021multi}) for comparison with FAST-LIO2\cite{xu2022fast} Faster-LIO\cite{bai2022faster} and Point-LIO\cite{he2023point}.
In addition, all the experiments are performed on a desktop computer equipped with an AMD Ryzen7 4800H CPU (8 cores, 2.9 G Hz) and 16G memory. 
In the following, we first briefly introduce the datasets we used. Subsequently, we verify the system's ability to handle aggressive motion on the Newer College dataset and our own dataset, and then conduct an ablation experiment to illustrate why our system tracks robustly under aggressive motion scene. In addition, We evaluate the accuracy and efficiency of the whole system on the NTU VIRAL, M2DGR, and UrbanNav datasets. Finally, we use the UrbanNav dataset again to further illustrate the high efficiency of the RC-Vox.

\begin{table}[htbp]
    \vspace{0.3cm}
  \centering
  \begin{threeparttable}
  \caption{Dataset Information}
    \begin{tabular}{cccc}
    \toprule
    \toprule
    \multirow{2}[4]{*}{Dataset} & \multicolumn{1}{c}{\multirow{2}[4]{*}{Sequence ID}} & \multicolumn{2}{c}{Sequence Infomation} \\
\cmidrule{3-4}    \multicolumn{1}{c}{} &       & \multicolumn{1}{c}{Length (m)} & \multicolumn{1}{c}{Duratioin (s)} \\
    \midrule
    \multirow{3}[2]{*}{NTU VIRAL} & \multicolumn{1}{c}{eee\_01-03} & 535   & 900 \\
    \multicolumn{1}{c}{} & \multicolumn{1}{c}{nya\_01-03} & 724   & 1235 \\
    \multicolumn{1}{c}{} & \multicolumn{1}{c}{sbs\_01-03} & 583   & 1116 \\
    \midrule
    M2DGR & \multicolumn{1}{c}{street\_01-10} & 7723  & 7809 \\
    \midrule
    \multirow{3}[2]{*}{UrbanNav} & \multicolumn{1}{c}{Mongkok} & 4860  & 3367 \\
    \multicolumn{1}{c}{} & \multicolumn{1}{c}{TST} & 3640  & 785 \\
    \multicolumn{1}{c}{} & \multicolumn{1}{c}{Whampoa} & 4510  & 1536 \\
    \midrule
    \multirow{2}[2]{*}{Newer College} & \multicolumn{1}{c}{\begin{tabular}[c]{@{}c@{}}06\_dynamic\_\\spinning\end{tabular}} & 97    & 119 \\
    \multicolumn{1}{c}{} & \multicolumn{1}{c}{\begin{tabular}[c]{@{}c@{}}collection\_1-\\quad-hard\end{tabular}} & 234   & 187 \\
    \midrule
    Ours  & \multicolumn{1}{c}{indoor\_01-03} & 259   & 308 \\
    \midrule
    \multicolumn{2}{c}{Total} & 23165 & 17362 \\
    \bottomrule
    \bottomrule
    \end{tabular}%
    \begin{tablenotes}    %这行要添加， 从这开始
        \footnotesize               %这行要添加
        \item[*] Angular velocities up to 3.3 rad/s in the Newer College dataset and 21.8 rad/s in our own dataset.
    \end{tablenotes}            %这行要添加
    \end{threeparttable}
  \label{tab:1}%
  \vspace{-0.3cm}
\end{table}%

\subsection{Dataset}
All these 4 public datasets are multi-sensor datasets, we only introduce the lidar and IMU among them. The NTU VIRAL dataset is collected with an unmanned aerial vehicle which has two 10 Hz Ouster OS1-16 Gen1 lidars (we only use the horizontal lidar) and a 385 Hz VectorNav VN100 IMU. The M2DGR dataset is collected with a ground robot which has a 10 Hz Velodyne VLP-32C lidar and a 150 Hz Handsfree A9 IMU. The UrbanNav dataset is collected with a human-driving robocar which has three 10 Hz lidars (we only used the Velodyne HDL-32E lidar among them) and a 400 Hz Xsens Mti 10 IMU. The Newer College dataset has two parts. The first part uses an Ouster OS1-64 Gen 1 lidar with a built-in 100Hz ICM-20948 IMU. The second part uses an Ouster OS0-128 lidar with a built-in 100Hz ICM-20948 IMU. All of these two parts are collected by handheld. And our own dataset is also collected by handheld with a Livox lidar Avia. The details of all sequences we used are listed in Table \ref{tab:1}.

\subsection{Aggressive Motion Test}
\begin{figure}[]       %不带*单栏，带*双栏
    \vspace{-3.2cm}
    \centering
    \includegraphics[scale=0.34]{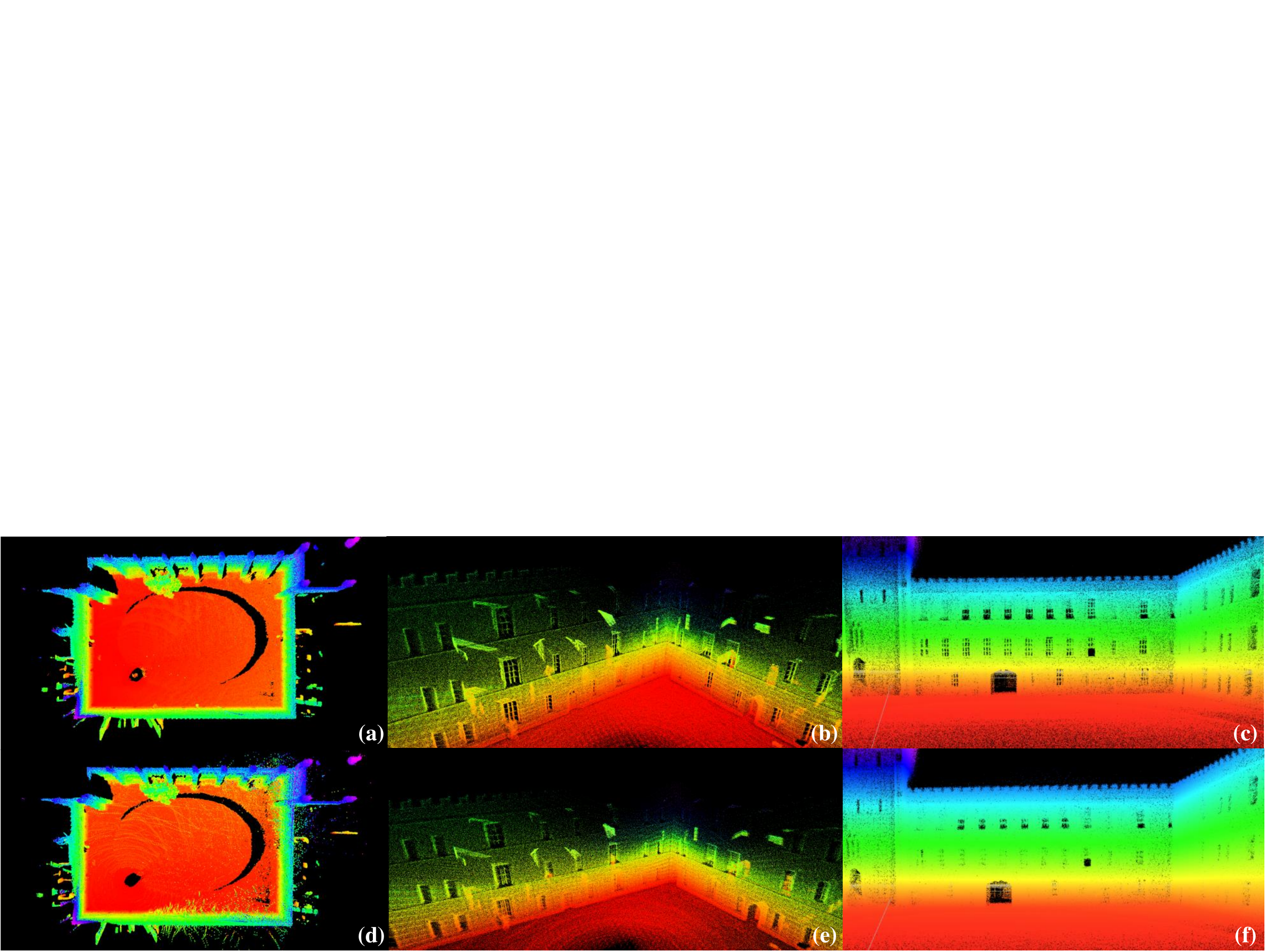}  %scale缩放比例，Fig.jpg文件名
    \caption{Point cloud reconstructed by our system and FAST-LIO2 on two aggressive motion sequences of the Newer College dataset. (a), (b) are reconstructed by our system on \emph{06\_dynamic\_spinning} and (c) is reconstructed by our system on \emph{collection\_1-quad-hard}. (d)-(f) are reconstructed by FAST-LIO2 and correspond to (a)-(c).}   % 图片名称
    \label{fig:4}
    \vspace{-0.1cm}
\end{figure}

\begin{figure}[]       %不带*单栏，带*双栏
    \vspace{0cm}
    \centering
    \includegraphics[scale=0.4]{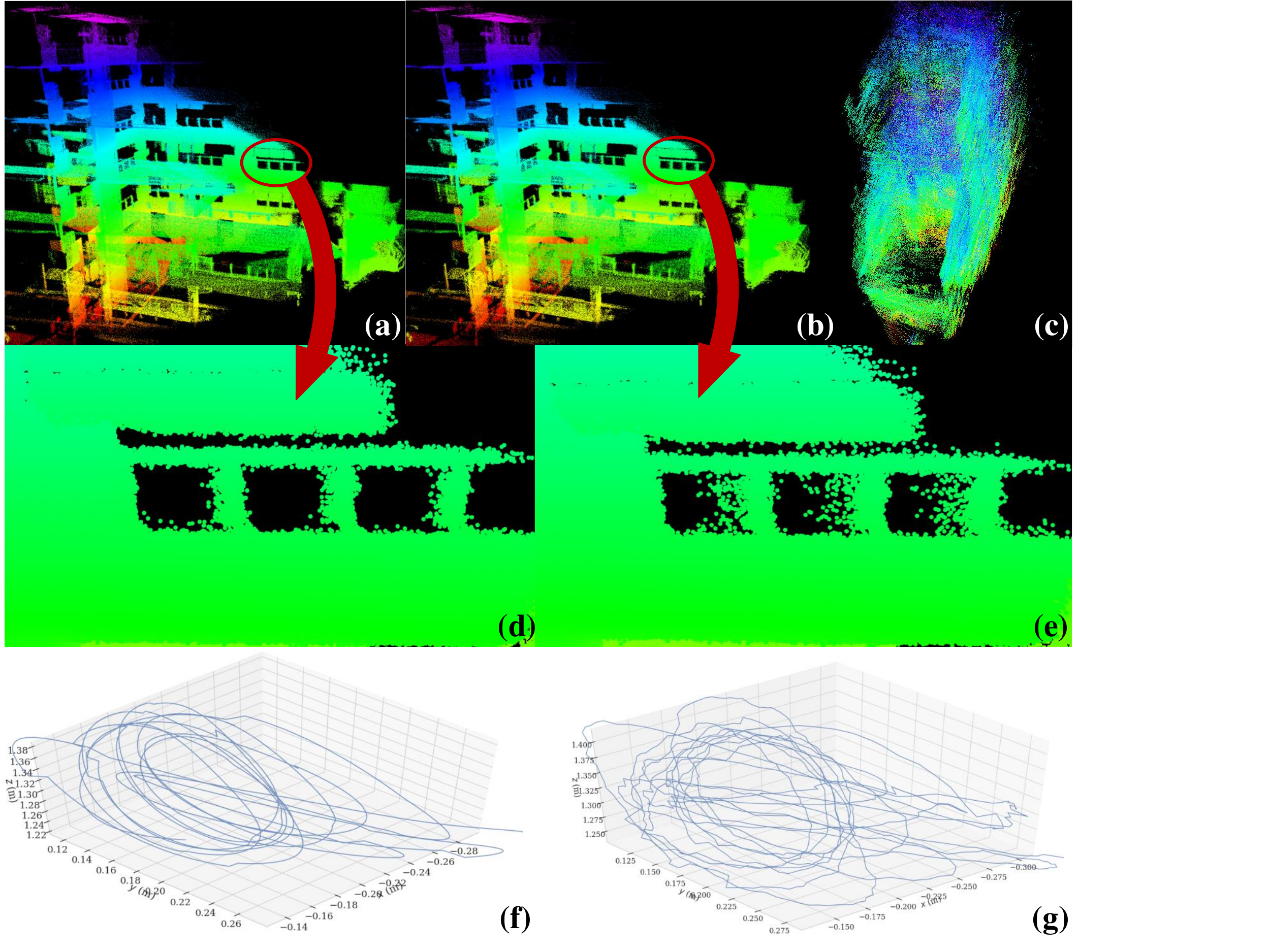}  %scale缩放比例，Fig.jpg文件名
    \caption{Ablation experiment on \emph{indoor\_01} sequence. (a)-(c) are the point cloud reconstructed by the full system, the system turning off backward smooth function and the system turning off sub-frame generation function respectively. (d) and (e) are the map details of the full system and the system turning off backward smooth function respectively. (f) and (g) are the trajectory segments of the full system and the system turning off backward smooth function respectively.}
    \label{fig:5}
    \vspace{-0.5cm}
\end{figure}

To verify the system's effectiveness in aggressive motion scence, qualitative tests are conducted on our own dataset (\emph{indoor\_01-03}) and two aggressive motion sequences of the Newer College dataset (\emph{06\_dynamic\_spinning} and \emph{collection\_1-quad-hard}). Meanwhile, we compare the result with FAST-LIO2. As shown in Fig. \ref{fig:1} and Fig. \ref{fig:4}, point cloud reconstructed by our system is clearer and shaper in the Newer College dataset. For our own dataset which has a more intense motion, FAST-LIO2 can no longer cope with it and fails to track, while our method still guarantees a high map accuracy.
To further explore the reason why our system can achieve better results, we conduct an ablation experiment on \emph{indoor\_01} sequence. In the experiment, we turn off the backward smooth function and the sub-frame generation function (i.e. without using the lidar points division) respectively. As shown in Fig. \ref{fig:5}, the lidar points division effectively reduces the accumulated error during IMU integration, thus ensuring stable tracking. In addition, we can notice that the overall mapping result still keeps a certain level of precision after turning off backward smooth, which is because partial constraint information still exists in the sub-frame and accurate alignment is still possible in the constraint direction. Therefore, the overall mapping performance is almost unaffected when the pose estimation error is tolerable in the degraded direction. However, as shown by the map details and the trajectory, the degraded direction causes ghosting in the map due to incorrect pose estimation and terrible track smoothness.

\begin{table}[htbp]
    \vspace{0.3cm}
  \centering
  \caption{Accuracy Test\tnote{*}}
  \begin{threeparttable}
    \begin{tabular}{ccccc}
      \toprule
      \toprule
      \multirow{2}[4]{*}{Sequence ID} & \multicolumn{1}{c}{\begin{tabular}[c]{@{}c@{}}FAST-LIO2\end{tabular}} & \multicolumn{1}{c}{\begin{tabular}[c]{@{}c@{}}Faster-LIO\end{tabular}} & \multicolumn{1}{c}{\begin{tabular}[c]{@{}c@{}}Point-LIO\\\end{tabular}} & \multicolumn{1}{c}{\begin{tabular}[c]{@{}c@{}}Ours\end{tabular}} \\
  \cmidrule{2-5}    \multicolumn{1}{c}{} & \multicolumn{4}{c}{\centering RMSE (m)} \\
      \midrule
      eee\_01 & 0.135 & 0.144 & 0.195 & \textcolor[rgb]{ 1,  0,  0}{0.132} \\
      eee\_02 & 0.118 & 0.121 & 0.130  & \textcolor[rgb]{ 1,  0,  0}{0.118} \\
      eee\_03 & 0.164 & \textcolor[rgb]{ 1,  0,  0}{0.162} & 0.182 & 0.164 \\
      nya\_01 & 0.121 & 0.132 & 0.182 & \textcolor[rgb]{ 1,  0,  0}{0.120} \\
      nya\_02 & 0.140  & 0.172 & 0.164 & \textcolor[rgb]{ 1,  0,  0}{0.139} \\
      nya\_03 & 0.150  & 0.154 & 0.190  & \textcolor[rgb]{ 1,  0,  0}{0.147} \\
      sbs\_01 & \textcolor[rgb]{ 1,  0,  0}{0.144} & \textcolor[rgb]{ 1,  0,  0}{0.144} & 0.197 & \textcolor[rgb]{ 1,  0,  0}{0.144} \\
      sbs\_02 & 0.144 & 0.144 & 0.214 & \textcolor[rgb]{ 1,  0,  0}{0.143} \\
      sbs\_03 & \textcolor[rgb]{ 1,  0,  0}{0.132} & 0.134 & 0.189 & \textcolor[rgb]{ 1,  0,  0}{0.132} \\
      \midrule
      street\_01 & 0.366 & 0.365 & 0.589 & \textcolor[rgb]{ 1,  0,  0}{0.318} \\
      street\_02 & 2.881 & 2.890  & 2.796 & \textcolor[rgb]{ 1,  0,  0}{2.580} \\
      street\_03 & 0.123 & 0.120  & \textcolor[rgb]{ 1,  0,  0}{0.110} & 0.114 \\
      street\_04 & 0.486 & 0.605 & \textcolor[rgb]{ 1,  0,  0}{0.409} & 0.571 \\
      street\_05 & 0.327 & 0.329 & 0.419 & \textcolor[rgb]{ 1,  0,  0}{0.213} \\
      street\_06 & 0.357 & 0.341 & 0.471 & \textcolor[rgb]{ 1,  0,  0}{0.307} \\
      street\_07 & 0.338 & 0.331 & 5.784 & \textcolor[rgb]{ 1,  0,  0}{0.254} \\
      street\_08 & 0.186 & 0.178 & 0.183 & \textcolor[rgb]{ 1,  0,  0}{0.176} \\
      street\_09 & 1.547 & \textcolor[rgb]{ 1,  0,  0}{1.523} & 1.942 & 1.575 \\
      street\_10 & 0.176 & 0.172 & 3.429 & \textcolor[rgb]{ 1,  0,  0}{0.171} \\
      \midrule
      Mongkok & 0.638 & 0.655 & \textcolor[rgb]{ 1,  0,  0}{0.470} & 0.657 \\
      TST   & 0.401 & 0.403 & 0.410  & \textcolor[rgb]{ 1,  0,  0}{0.380} \\
      Whampoa & 0.255 & 0.254 & 0.280  & \textcolor[rgb]{ 1,  0,  0}{0.245} \\
      \bottomrule
      \bottomrule
      \end{tabular}%
    \begin{tablenotes}    %这行要添加， 从这开始
        \footnotesize               %这行要添加
        \item[*] The NTU VIRAL and M2DGR datasets are evaluated using the ATE metric, while the UrbanNav dataset is evaluated using the RTE metric (per 10 meters) because of the longer trajectory and larger drift error.          %这行要添加
        % \item[2,3] The quick brown fox jumps over the lazy dog.        %这行要添加
    \end{tablenotes}            %这行要添加
  \label{tab:2}%
  \end{threeparttable}
\end{table}%

\begin{table}[htbp]
  \centering
  \caption{Efficiency Test}
  \begin{threeparttable}
    \begin{tabular}{cccccc}
    \toprule
    \toprule
    \multirow{2}[4]{*}{Sequence ID} & \multicolumn{1}{c}{\begin{tabular}[c]{@{}c@{}}FAST-\\LIO2\\(Single)\end{tabular}} & \multicolumn{1}{c}{\begin{tabular}[c]{@{}c@{}}FAST-\\LIO2\\(Paral)\end{tabular}} & \multicolumn{1}{c}{\begin{tabular}[c]{@{}c@{}}Faster\\-LIO\\(Single)\end{tabular}} & \multicolumn{1}{c}{\begin{tabular}[c]{@{}c@{}}Faster\\-LIO\\(Paral)\end{tabular}} & \multicolumn{1}{c}{\begin{tabular}[c]{@{}c@{}}Ours\tnote{*}\\(Single)\end{tabular}} \\
\cmidrule{2-6}    \multicolumn{1}{c}{} & \multicolumn{5}{c}{Time Cost Per Frame (ms)} \\
    \midrule
    eee\_01 & 30.5  & 17.5  & 47.0    & 10.9  & \textcolor[rgb]{ 1,  0,  0}{9.3} \\
    eee\_02 & 27.7  & 17.3  & 41.1  & 11.1  & \textcolor[rgb]{ 1,  0,  0}{8.9} \\
    eee\_03 & 28.2  & 17.0    & 37.2  & 10.7  & \textcolor[rgb]{ 1,  0,  0}{8.7} \\
    nya\_01 & 23.7  & 15.7  & 45.7  & 11.5  & \textcolor[rgb]{ 1,  0,  0}{7.1} \\
    nya\_02 & 24.9  & 15.9  & 49.7  & 12.0    & \textcolor[rgb]{ 1,  0,  0}{7.6} \\
    nya\_03 & 23.1  & 15.7  & 50.4  & 12.1  & \textcolor[rgb]{ 1,  0,  0}{7.4} \\
    sbs\_01 & 23.5  & 15.8  & 35.8  & 10.1  & \textcolor[rgb]{ 1,  0,  0}{7.9} \\
    sbs\_02 & 23.9  & 15.8  & 40.8  & 10.9  & \textcolor[rgb]{ 1,  0,  0}{8.1} \\
    sbs\_03 & 24.0    & 16.1  & 41.1  & 10.5  & \textcolor[rgb]{ 1,  0,  0}{7.6} \\
    \midrule
    street\_01 & 45.0    & 25.5  & 68.5  & 16.5  & \textcolor[rgb]{ 1,  0,  0}{13.2} \\
    street\_02 & 39.4  & 23.2  & 53.9  & 14.8  & \textcolor[rgb]{ 1,  0,  0}{12.2} \\
    street\_03 & 52.0    & 28.5  & 92.7  & 19.4  & \textcolor[rgb]{ 1,  0,  0}{14.1} \\
    street\_04 & 42.4  & 25.4  & 58.1  & 15.5  & \textcolor[rgb]{ 1,  0,  0}{12.7} \\
    street\_05 & 36.7  & 22.8  & 51.6  & 14.7  & \textcolor[rgb]{ 1,  0,  0}{11.8} \\
    street\_06 & 39.8  & 24.5  & 54.8  & 15.5  & \textcolor[rgb]{ 1,  0,  0}{12.5} \\
    street\_07 & 45.5  & 26.8  & 69.6  & 17.0    & \textcolor[rgb]{ 1,  0,  0}{12.7} \\
    street\_08 & 49.8  & 29.0    & 89.3  & 19.8  & \textcolor[rgb]{ 1,  0,  0}{13.8} \\
    street\_09 & 44.1  & 26.5  & 64.2  & 16.7  & \textcolor[rgb]{ 1,  0,  0}{13.0} \\
    street\_10 & 39.3  & 25.3  & 57.1  & 16.1  & \textcolor[rgb]{ 1,  0,  0}{11.2} \\
    \midrule
    Mongkok & 64.2 & 21.5  & 48.9  & 12.7  & \textcolor[rgb]{ 1,  0,  0}{8.3} \\
    TST   & 83.3 & 26.5  & 43.1  & 12.9  & \textcolor[rgb]{ 1,  0,  0}{9.8} \\
    Whampoa & 73.3 & 23.8  & 42.4  & 12.7  & \textcolor[rgb]{ 1,  0,  0}{8.6} \\
    \bottomrule
    \bottomrule
    \end{tabular}%
    \begin{tablenotes}    %这行要添加， 从这开始
        \footnotesize               %这行要添加
        \item[*] For fairness, the time consumption of our system is not the processing time of a single sub-frame, but the time to process the full scan (including sub-frame generation, processing multiple sub-frames and mapping).
    \end{tablenotes}            %这行要添加
  \label{tab:3}%
  \end{threeparttable}
\end{table}%

\subsection{System Accuracy and Efficiency Test}
We use a total of 22 sequences for system accuracy and efficiency evaluation. In addition, we also test the performance of FAST-LIO2 and Faster-LIO with or without parallelized acceleration in the efficiency test, the parallelized acceleration version is represented by \emph{(Paral)} and the single-thread version is represented by \emph{(Single)}. In the following experiments, FAST-LIO2, Faster-LIO, and our system all apply a 4-point skip operation and a 0.5m voxel filter to the original point cloud. 

Table \ref{tab:2} shows that our system achieves the same or higher accuracy compared with other algorithms.Especially in \emph{street\_07} sequence, the robot performs a lot of rotations, and our system effectively improves the system accuracy through lidar points division and the ESKS framework. However, in \emph{street\_04} sequence and \emph{Mongkok} sequence, we can see that the accuracy of our system and Faster-LIO are lower than FAST-LIO2, which is because that the k-d tree based LIO method has a certain accuracy advantage over the voxel based method. 

In terms of efficiency (as shown in Table \ref{tab:3}), we can see that the single-thread version of FAST-LIO2 is slightly faster than Faster-LIO, but Faster-LIO is more suitable for parallelized acceleration. Notably, the single-thread version of our system is even faster than the parallelized version of Faster-LIO. We also give the time consumption of each module of our system in the UrbanNav dataset (as shown in Table \ref{tab:4}). 

\begin{table}[]
    \vspace{0.3cm}
  \centering
  \caption{Time consumption for each module of our system}
    \begin{tabular}{c|ccc}
    \toprule
    \toprule
    Steps & \multicolumn{1}{c}{\begin{tabular}[c]{@{}c@{}}Mongkok\\(ms)\end{tabular}} & \multicolumn{1}{c}{\begin{tabular}[c]{@{}c@{}}TST\\(ms)\end{tabular}} & \multicolumn{1}{c}{\begin{tabular}[c]{@{}c@{}}Whampoa\\(ms)\end{tabular}} \\
    \midrule
    Sub-frame generation & 1.04 & 1.11 & 1.02 \\
    \midrule
    \begin{tabular}[c]{@{}c@{}}Propagation + Update\end{tabular} & 6.25 & 7.09 & 6.40 \\
    \midrule
    \begin{tabular}[c]{@{}c@{}}Backward smooth + \\Constraint integrity analysis\end{tabular}  & 0.53 & 0.65 & 0.52 \\
    \midrule
    \begin{tabular}[c]{@{}c@{}}Incremental mapping\\ + Map delete\end{tabular} & 0.47 & 1.02 & 0.64 \\
    \midrule
    Total runtime & 8.31 & 9.88 & 8.60 \\
    \bottomrule
    \bottomrule
    \end{tabular}%
  \label{tab:4}%
\end{table}%
\begin{figure}[]       %不带*单栏，带*双栏
    \centering
    \includegraphics[scale=0.34]{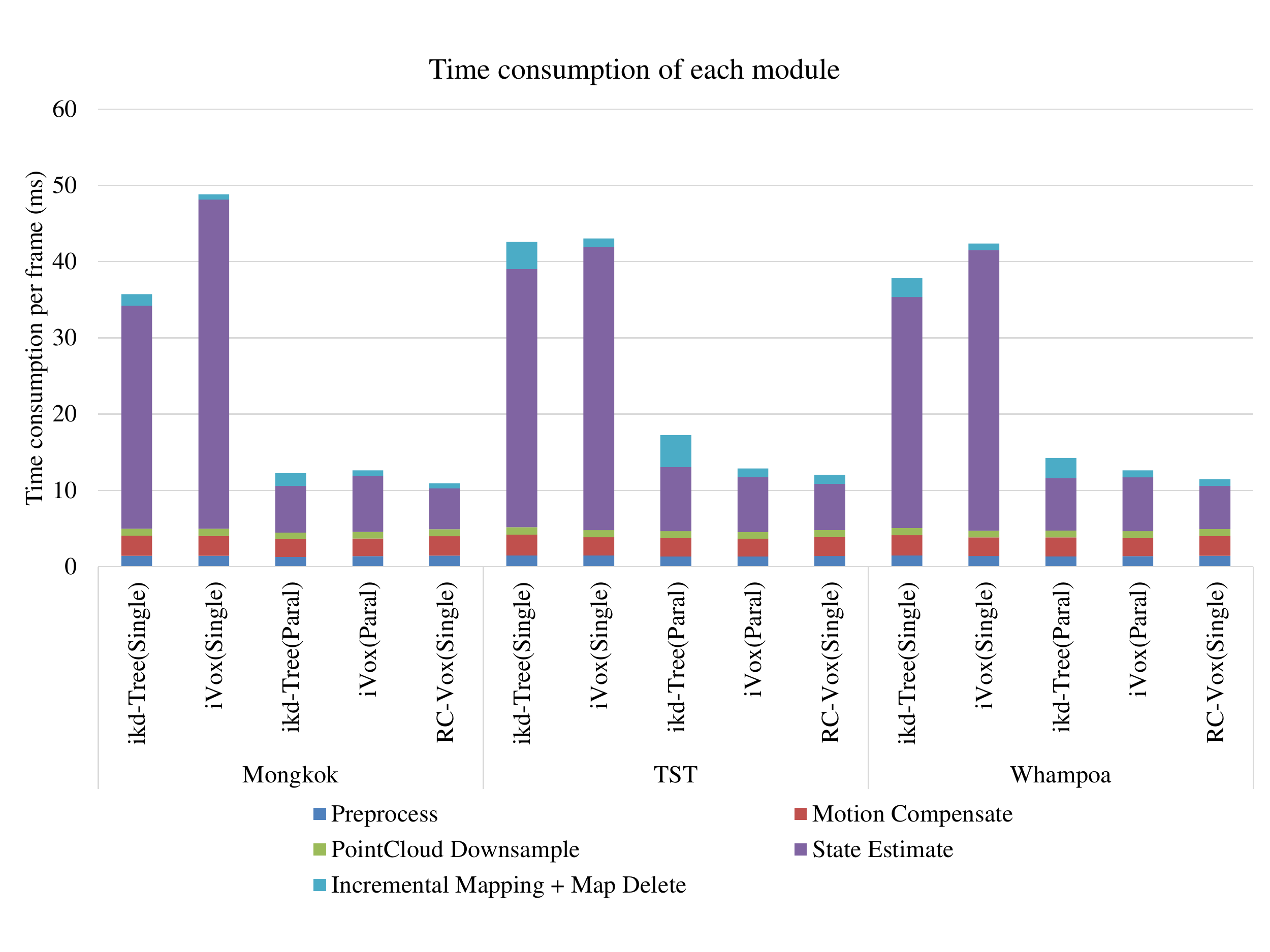}  %scale缩放比例，Fig.jpg文件名
    \caption{Speed test of ikd-Tree, iVox and RC-Vox}   % 图片名称
    \label{fig:6}
\end{figure}
\subsection{RC-Vox Speed Test}
To further compare the efficiency of RC-Vox, ikd-Tree, and iVox, we modify Faster-LIO with its map part to RC-Vox and ikd-Tree respectively. The time consumption of each module is shown in Fig. \ref{fig:6}. For the incremental mapping and map delete phases, \emph{ikd-Tree(Single)} takes the longest time. Although \emph{RC-Vox(Single)} needs to record points into nearby voxels, thanks to the array structure which allows for fast access, it can achieve about the same time consumption as \emph{iVox(Paral)}. For the state estimation phase, \emph{RC-Vox(Single)} has the shortest time consumption due to the array structure and our optimization of the k-NN search. Moreover, it can be seen that \emph{RC-Vox(Single)} is even faster than \emph{ikd-Tree(Paral)} and \emph{iVox(Paral)}. All the above shows the superiority of the RC-Vox in terms of speed.

\section{CONCLUSIONS}
This paper proposes a fast and robust LIO system that can stably cope with aggressive motion while ensuring tracking accuracy in general scenes. For scenes with aggressive motion, we effectively reduce the IMU integration error through an adaptive division of the lidar points. To cope with the possible degradation of the sub-frames due to insufficient constraints, we propose a tightly coupled iterated ESKS framework to conduct the pose estimation. Regarding system efficiency, we propose a robocentric voxel map structure RC-Vox, based on a fixed-size two-layer 3D array structure. Using a modulo operation, RC-Vox enables efficient local map maintenance by mapping local map points into this fixed-size array structure. In addition, we further optimize the k-NN search by moving the search operation of nearby map points from the state estimation phase to the mapping phase, thus further improving the system's efficiency.

\addtolength{\textheight}{-12cm}   % This command serves to balance the column lengths
\bibliography{ref}

\end{document}